\documentclass[
]{ceurart}

\usepackage{listings}
\begin{document}

\copyrightyear{2023}
\copyrightclause{Copyright for this paper by its authors. Use permitted under Creative Commons License Attribution 4.0 International (CC BY 4.0).\\Editors: Mohammad Aliannejadi, Guglielmo Faggioli, Nicola Ferro, Michalis Vlachos
}

\conference{CLEF 2023: Conference and Labs of the Evaluation Forum, September 18–21, 2023, Thessaloniki, Greece}

\title{SINAI at eRisk@CLEF 2023: Approaching Early Detection of Gambling with Natural Language Processing}

\title[mode=sub]{Notebook for the Early Risk Prediction on the Internet Lab at CLEF 2023}


\author[1]{Alba María Mármol-Romero}[%
orcid=0000-0001-7952-4541,
email=amarmol@ujaen.es,
]
\author[2]{Flor Miriam Plaza-del-Arco}[%
orcid=0000-0002-3020-5512,
email=flor.plaza@unibocconi.it,
]
\author[1]{Arturo Montejo-Ráez}[%
orcid=0000-0002-8643-2714,
email=amontejo@ujaen.es,
]
\address[1]{Computer Science Department, SINAI, CEATIC, Universidad de Jaén, 23071, Spain}
\address[2]{Bocconi University, Via Sarfatti 25, 20100, Milan, Italy}

\begin{abstract}
  This paper describes the participation of the SINAI team in the eRisk@CLEF lab. Specifically, one of the proposed tasks has been addressed: Task 2 on the early detection of signs of pathological gambling. The approach presented in Task 2 is based on pre-trained models from Transformers architecture with comprehensive preprocessing data and data balancing techniques. Moreover, we integrate Long-short Term Memory (LSTM) architecture with automodels from Transformers. In this Task, our team has been ranked in seventh position, with an F1 score of 0.126, out of 49 participant submissions and achieves the highest values in recall metrics and metrics related to early detection.
\end{abstract}


\maketitle

\section{Introduction}

The large amount of content posted daily on social media has made them a significant source of data for the early detection of mental disorders and risky behaviours. The eRisk@CLEF 2023 lab \cite{eRisk20232} focuses on early risk prediction on the Internet and its goal is to promote the development of automatic systems for the detection of mental disorders such as depression, self-harm or eating disorders. In this edition, three tasks have been proposed:

\begin{itemize}
    \item Task 1: Search for symptoms of depression. It is a new task consisting of ranking sentences from a collection of user writings according to their relevance to a depression symptom. Then, the participants will have to provide rankings for the 21 symptoms of depression from the BDI Questionnaire
    \item Task 2: Early Detection of Signs of Pathological Gambling. It involves sequentially processing writings and detecting as early as possible the first signs of pathological gambling. It is a continuation of Task 1 proposed for eRisk 2021 and for eRisk 2022, but the difference is that in this edition a larger amount of training data has been provided. The training data of this edition comprises all test users of the 2021 and 2022 tasks.
    \item Task 3: Measuring the severity of the signs of Eating Disorders. Its aim is to estimate a user's level of disordered eating from his or her history of posts. For this purpose, for each user, a standard eating disorder questionnaire (EDE-Q) has to be filled in.
\end{itemize}

Currently, our research group SINAI\footnote{\url{https://sinai.ujaen.es}} is working on the PRECOM project\footnote{\url{https://precom.ujaen.es/}} focused on the early detection of gambling addiction risk behaviour. Therefore, our interest in developing systems as those expected to answer eRisk tasks is high, as it is a perfect playground to test our approaches.

In this sense, our main goals are not only to produce systems reporting high performance but to understand the best methods and approaches that can be applied in similar scenarios. It is not our aim to put as many features and as many systems as possible all together in an ensemble of predictors to gain the top ranking position, but rather to find out the best approaches that can be applied to our project's objectives. The design of online and monitoring tools, as is requested in Task 1, along with the ability to understand user's disorder, the main pursuit in Task 3, fully matches our research interests.

This work presents the participation of our research group, the SINAI team, in \textit{Task 2: Early Detection of Signs of Pathological Gambling}.

The rest of the paper is organized as follows. Section 2 describes the details of our participation in Task 2. It is divided into subsections in which, first, we introduce what the task consists of, the data provided and the evaluation measures used. Secondly, the system developed and the methodology used is presented. Thirdly, the experimental setup is detailed. Subsequently, the results obtained and a discussion of them are presented. Finally, Section 3 shows the conclusions obtained after participation in the eRisk lab and the perspectives for future work.

\section{Task 2: Early Detection of Signs of Pathological Gambling}

\subsection{Task description}

This particular task focuses on identifying signs of gambling addiction at an early stage by analyzing posts from social media in the exact order they were published. The participating systems were required to read the posts, process them, and generate a response to proceed to the next set of posts. The dataset used for this task consists of 33,719 posts from 103 individuals who were categorized as having a positive association with gambling addiction, as well as approximately 1 million posts from 2,071 individuals who were not categorized as addicts \cite{erisk2023}.

The task is approached from two different perspectives: as a binary decision problem and as a ranking (regression) decision problem. In the binary decision problem, the posts need to be classified as either positive (label 1, indicating the presence of addiction) or negative (label 0, indicating no addiction). The earlier the system detects the presence of addiction, the better it performs, as reflected by the evaluation metrics proposed by the organizers, namely ERDE and $F_{latency}$, along with well-known measures such as precision, recall, and F1 scores. In the ranking decision problem, instead of assigning binary labels, the system computes a score representing the estimated risk of developing a gambling addiction. Various metrics commonly used in information retrieval, such as P@10 or NDCG, are employed to evaluate this alternative perspective of the task.

\subsection{System and methods}
To tackle this task, we adopted a supervised learning approach. Our models were trained using the provided training dataset from the eRisk organizers, which combined data from the years 2021 and 2022. This dataset consists of a time series of posts authored by various users. Our main contributions in this task have been the data processing work done as well as the handling of the unbalanced data and the development of a new system that integrates a transformer with an LSTM network.

We performed two different methods of pre-processing on the training data: one comprehensive method and another lighter method tailored to the specific system structure where it is employed. The lighter method simply replaces URLs with tags, while the comprehensive method involves a series of steps. Here is a detailed explanation of the extensive processing we conducted:

\begin{itemize}
\item We replaced all numeric HTML entities in the post string with their corresponding Unicode character equivalents.
\item Users, URLs, and emails mentioned in the messages were identified and replaced with \#USER, \#URL, and \#EMAIL, respectively.
\item We removed any empty characters and replaced special Unicode white-space characters that may have been present in the text.
\item Emojis were substituted with strings using the emoji library\footnote{https://pypi.org/project/emoji/}.
\item If more than seven identical punctuation marks appeared consecutively, they were reduced to only three instances.
\item Markdown formatting was removed from the text.
\item We standardized the use of double and single quotes in the text string, replacing curved quotes and curved single quotes with straight double and single quotes, respectively.
\end{itemize}

\begin{figure}[!h]
\centering
\includegraphics[width=1\textwidth]{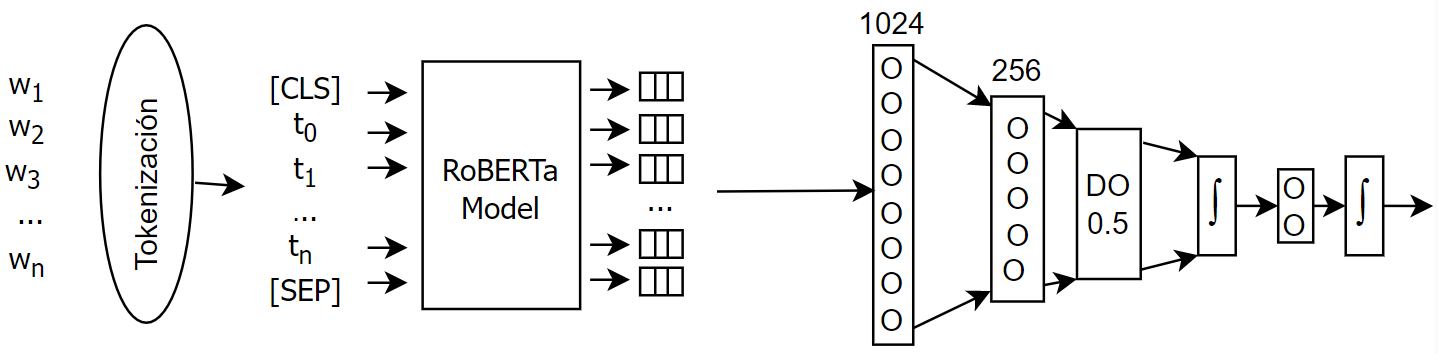}
\caption{Model architecture for the binary decision task involving the RoBERTa-Large model or the XLM-RoBERTa Large model. The input representation is based on RoBERTa-Large tokenization or XLM-RoBERTa Large tokenization. Each embedding of the model output is passed through a feed-forward neural network with a hidden layer, generating a binary decision and a score as the logits.}
\label{model1}
\end{figure}

For our participation in eRisk Task 2, we trained a total of five models. Two systems utilized the XLM-Roberta-Large \cite{xlmroberta} model, while the other three employed the RoBERTa-Large \cite{roberta} linguistic model. Four of these systems followed the architecture depicted in Figure \ref{model1}.

In these four systems, although the models used and the datasets differed, they followed the same structure. The prediction process for Task 2 involved post extraction. All posts by each user were concatenated, and if the concatenated posts exceeded 500 words, the subject was split into sub-subjects. This ensured that the maximum word limit was not exceeded. The concatenated posts served as the input to either the RoBERTa-Large automodel or the XLM-RoBERTa Large automodel. Subsequently, a Feed Forward Neural Network (FFNN) with a single hidden layer was applied to generate the final predictions. The initial layer of the FFNN had an input dimension of 1024 from the embeddings vector generated by RoBERTa. The output size of this layer was set to 256. During the learning process, these outputs were passed through a dropout layer with a probability of 0.5. Finally, a ReLU activation function was applied before feeding the last FFNN with 256 inputs and 2 outputs.

\begin{figure}[!h]
\centering
\includegraphics[width=1\textwidth]{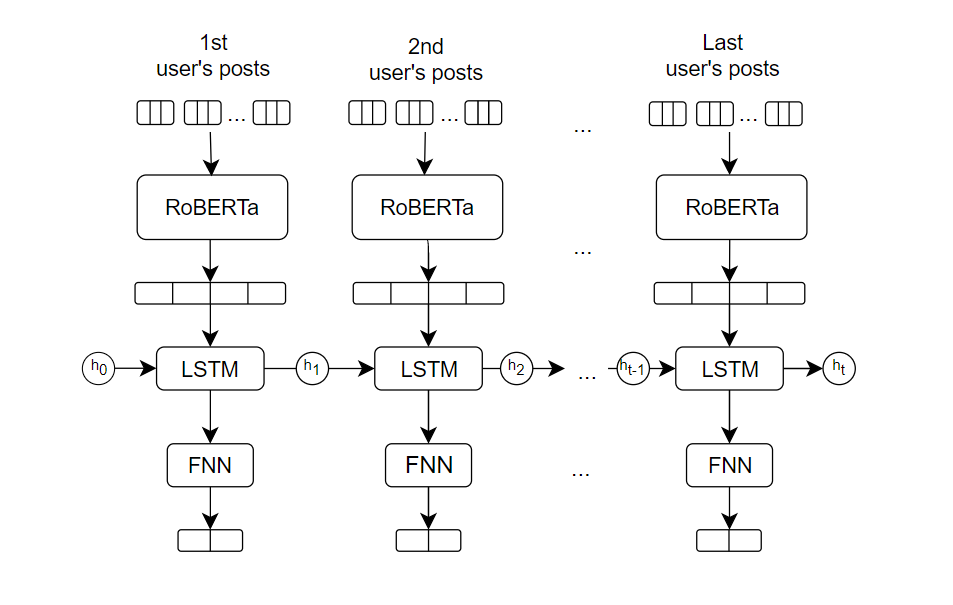}
\caption{Model architecture for the binary decision task involving the RoBERTa model and LSTM architecture. The input representation consists of RoBERTa-Large tokenization of each subject's message in each round. Each embedding of the model output is passed through an LSTM layer. The output of the LSTM is then fed into a feedforward neural network with one hidden layer, producing a binary decision. }
\label{model2}
\end{figure}

However, the most innovative system among the others is the one that combines the RoBERTa-Large model with the LSTM architecture, as illustrated in Figure \ref{model2}.

Regarding the system that integrates RoBERTa-Large and the LSTM architecture, the process begins by concatenating the last 50 posts from each subject. This is because we have studied this number is the most appropriate according to this type of dataset as we saw in our last year participation \cite{marmol2022sinai}. This concatenation is subsequently split into rounds, representing the input to the RoBERTa-Large automodel. This approach allows us to simulate the sequencing that will occur during the testing phase. The model's output is then passed to an LSTM layer. The output of the LSTM layer is further processed by a Feed Forward Neural Network (FFNN) with a single hidden layer, generating the final predictions. The initial layer of the FFNN has an input dimension of 1024, and the output of this layer is of size 256. These outputs are subjected to a dropout layer (with a probability of 0.5) during the learning process. Finally, a ReLU activation function is applied before feeding the last FFNN with 256 inputs and 2 outputs. Moreover, the cell state and hidden state are passed to the next round in each batch.

The systems were implemented using Python packages such as scikit-learn \cite{scikit}, Transformers \cite{transformers}, and PyTorch \cite{pytorch}. 

The training was performed on a 2xGPU NVIDIA V100 server.

\subsection{Experimental setup}
First, all the parameters established in these systems and presented in this document are the result of a previous hyperparameter optimisation search carried out with Optuna\footnote{https://optuna.org/}, including the dataset selected for data balancing. This means that given a dataset with a number of subjects, these subjects were divided into further subjects due to the sequence length limit given by the Transformers models. For this new set, only a part of the subjects belonging to the control group was chosen due to the initial imbalance of the data. Table \ref{tab:optuna}, shows the information concerning the search space. However, for the latest system that includes the LSTM (run 4), the parameters were chosen according to what our hardware architecture and training time limit allowed.

\begin{table}[htp!]
    \caption{Search space in the hyperparameter optimisation process. In this case, with the variable "Data" we refer to the number of sub-subjects belonging to the control group we are working with (in the order from the first generated to the last).}
    \centering
        \begin{tabular}{c|c}
         \hline
         \textbf{Parameter} & \textbf{Search space} \\
         \hline
         Learning rate & [1e-4, 1e-5, 1e-6]\\
         Batch size & [8, 16]\\
         Epoch & [1, 2, 3, 4]\\
         Weight decay & [0.1, 0.01, 0.001]\\
         Max length & [256, 512]\\
         Data & [5, 10, 15, 20, All]\\
         \hline
    \end{tabular}
    \label{tab:optuna}
\end{table}

After the search for hyperparameters, in the learning phase, a maximum length of 512 tokens per document was set. For runs 0 to 3, the document is the concatenation of an undetermined number of posts whose sum of their words is no bigger than 500 and the optimizer used was AdamW. For run four, the document is the concatenation of the last 50 posts of each subject and the optimizer used was also AdamW. In table \ref{tab:parameters} there is a summary of hyperparameters and models used in each run.

\begin{table}[htp!]
    \caption{Summary of hyperparameters and models used in each run}
    \centering
        \begin{tabular}{c|c|c|c|c|c}
         \hline
         \textbf{Run} & \textbf{Batch size}& \textbf{Learning rate} & \textbf{Weight decay} & \textbf{Epochs} & \textbf{Model} \\
         \hline
         Run 0 & 8 & 1e-5 & 0.01 & 3 & RoBERTa-Large\\
         Run 1 & 8 & 1e-5 & 0.01 & 2 & RoBERTa-Large\\
         Run 2 & 8 & 1e-5 & 0.01 & 3 & XLM-RoBERTa-Large\\
         Run 3 & 8 & 1e-5 & 0.01 & 2 & XLM-RoBERTa-Large\\
         Run 4 & 8 & 1e-6 &  0 & 7 & RoBERTa-Large \& LSTM\\
         \hline
    \end{tabular}
    \label{tab:parameters}
\end{table}

\begin{itemize}
    \item \textbf{Run 0}. This run utilized a binary classification model based on RoBERTa-Large. The document used for training was the concatenation of an undetermined number of posts, where the sum of their words did not exceed 500. However, due to the unbalanced nature of the provided data, only the first 15 sub-subjects generated by our preprocessing, with no risk of gambling pathology, and all the sub-subjects generated with the risk of gambling pathology were considered. The training dataset consisted of 14.78\% positive subjects and 85.22\% negative subjects. The model's output score determined whether a user was considered a potential gambler or not. Users with a score lower than 0.5 were not considered potential gamblers, while those with a score equal to or higher than 0.5 were classified as potential gamblers.
    \item \textbf{Run 1}. Similar to Run 0, this run employed a binary classification model based on RoBERTa-Large. The training dataset used the same structure as the other systems, considering the first 10 sub-subjects generated by our preprocessing with no risk of suffering gambling pathology, and all the sub-subjects generated with the risk of gambling pathology. The training dataset for this run contained 18.84\% positive subjects and 81.16\% negative subjects.
    \item \textbf{Run 2}. In this run, the model used was XLM-RoBERTa-Large, while maintaining the other settings from Run 0.
    \item \textbf{Run 3}. Similar to Run 1, this run utilized the XLM-RoBERTa-Large model instead of RoBERTa-Large, while keeping the remaining configurations the same.
    \item \textbf{Run 4}. This run involved a binary classification model based on RoBERTa-Large and the LSTM architecture. The document used for training differed from the other systems due to the longer training time required. Only the last 50 concatenated posts were considered, with the only preprocessing step being the replacement of URLs with \#URL. Additionally, 500 negative subjects were randomly selected from the unbalanced training dataset. The training dataset for this run comprised 245 positive subjects (the sum of positive subjects from the previous year's data) and 500 negative subjects. The model's output score determined whether a user was considered a potential gambler or not. Users with a score lower than 0.5 were not considered potential gamblers, while those with a score equal to or higher than 0.5 were classified as potential gamblers.
    \end{itemize}

\subsection{Results and discussion}

Salient results have been achieved with the approaches explored by our team. From the reported results provided by the organizers, we have extracted our scores, which are shown in Tables \ref{tab:task1_results1} and \ref{tab:task1_results2}.

\begin{table}[htp!]
    \caption{Results of SINAI team for Task 2 in decision-based evaluation}
    \centering
    \begin{tabular}{c|cccccccc}
        Run & $P$ & $R$ & $F1$ & $ERDE_5$ & $ERDE_{50}$ & $latency_{tp}$ & $speed$ & $latency_w F1$ \\ 
        \hline
        0 & 0.115 & 1.000 & 0.206 & 0.029 & 0.021 & 1.0 & 1.000 & 0.206 \\
        1 & 0.124 & 1.000 & 0.221 & 0.028 & 0.020 & 2.0 & 0.996 & 0.220 \\
        2 & 0.108 & 1.000 & 0.195 & 0.030 & 0.022 & 1.0 & 1.000 & 0.195 \\
        3 & 0.126 & 1.000 & 0.224 & 0.029 & 0.020 & 2.0 & 0.996  & 0.223 \\
        4 & 0.092 & 0.981 & 0.168 & 0.044 & 0.027 & 3.0 & 0.992 & 0.166 \\
    \end{tabular}
    \label{tab:task1_results1}
\end{table}

\begin{table}[htp!]
    \caption{Results of SINAI team for Task 2 in ranking-based evaluation}
    \centering
    \begin{tabular}{c|ccc|ccc|ccc}
    &   \multicolumn{3}{c}{1 writing} & \multicolumn{3}{c}{100 writing} & \multicolumn{3}{c}{500 writing} \\ \hline
        Run & \rotatebox[origin=c]{90}{$P@10$} & \rotatebox[origin=c]{90}{$NDCG@10$} & \rotatebox[origin=c]{90}{$NDCG@100$} & \rotatebox[origin=c]{90}{$P@10$} & \rotatebox[origin=c]{90}{$NDCG@10$} & \rotatebox[origin=c]{90}{$NDCG@100$} & \rotatebox[origin=c]{90}{$P@10$} & \rotatebox[origin=c]{90}{$NDCG@10$} & \rotatebox[origin=c]{90}{$NDCG@100$}  \\ 
        \hline
        0 & \textbf{1.00} & \textbf{1.00}  & 0.72 &\textbf{1.00}  & \textbf{1.00}  & 0.88 & \textbf{1.00}  & \textbf{1.00}  & 0.85\\ 
        1 & \textbf{1.00}  & \textbf{1.00}  & \textbf{0.73} & \textbf{1.00}  & \textbf{1.00}  & 0.90 & \textbf{1.00}  & \textbf{1.00}  & 0.85\\ 
        2 & \textbf{1.00}  & \textbf{1.00}  & 0.71 & \textbf{1.00}  & \textbf{1.00}  & 0.87 & \textbf{1.00}  & \textbf{1.00} & 0.84\\ 
        3 & \textbf{1.00}  & \textbf{1.00}  & 0.72 & \textbf{1.00}  & \textbf{1.00}  & 0.89 & \textbf{1.00}  & \textbf{1.00}  & 0.86\\ 
        4 & 0.88 & 0.86 & 0.53 & 0.90 & 0.94 & 0.56 & 0.70 & 0.80 & 0.47\\ 
    \end{tabular}
    \label{tab:task1_results2}
\end{table}

In decision-based evaluation, the value of the Recall score obtained by our Runs 0, 1, 2 and 3 (1.000) is the highest among all submissions by participants (49 submissions were reported in total). Our ERDE values are the second and fifteenth highest (0.020 and 0.028). In terms of $speed$ and $latency_{TP}$, we also reach top values for run 0 and run 2. Thus, it seems that a larger unbalance of the data and a larger number of epochs contribute slightly to the improvement in prediction. In general, the systems developed demonstrate a high capacity to correctly identify the majority of positive subjects at a fast speed.

Regarding ranking-based evaluation (see Table~\ref{tab:task1_results2}), our team reaches one of the highest positions. High NDCG and P@k values indicate that the systems are capable of providing relevant and quality recommendations. 

The differences in the systems developed according to the metrics appear to be minimal. This makes sense since the systems have almost the same structure. The only different system presented is used in run 4, which has slightly lower values than the rest of the systems, which can be understood by the resource limitations that have been taken into account due to the time required for execution. However, it is the one that we believe has the most promising results with the most research time invested.

\section{Conclusions and future work}

This paper describes our participation as SINAI team in Task 2 of the eRisk@CLEF 2023 edition. The former is the continuation of the first edition in 2021 and the second edition in 2022 and aims to detect signs of pathological gambling as soon as possible. For Task 2, we have developed only classification models using state-of-the-art pre-trained language models based on Transformers. Besides, we explored LSTM architecture's integration with this model type. 

In future work, we plan to analyze in depth the use of the LSTM architecture for this kind of problem as the main objective is early detection in a sequential way. We plan to test different training strategies with the integration of a transformers model together with the LSTM architecture. In addition, we expect to continue to work on data processing and the imbalance present in the data.

\begin{acknowledgments}
  This work has been partially supported by WeLee project (1380939, FEDER Andalucía 2014-2020) funded by the Andalusian Regional Government, and projects CONSENSO (PID2021-122263OB-C21), MODERATES (TED2021-130145B-I00), SocialTOX (PDC2022-133146-C21) funded by Plan Nacional I+D+i from the Spanish Government, and project PRECOM (SUBV-00016) funded by the Ministry of Consumer Affairs of the Spanish Government.
\end{acknowledgments}

\bibliography{sample-ceur}

\begin{thebibliography}{8}
\expandafter\ifx\csname natexlab\endcsname\relax\def\natexlab#1{#1}\fi
\providecommand{\url}[1]{\texttt{#1}}
\providecommand{\href}[2]{#2}
\providecommand{\path}[1]{#1}
\providecommand{\DOIprefix}{doi:}
\providecommand{\ArXivprefix}{arXiv:}
\providecommand{\URLprefix}{URL: }
\providecommand{\Pubmedprefix}{pmid:}
\providecommand{\doi}[1]{\href{http://dx.doi.org/#1}{\path{#1}}}
\providecommand{\Pubmed}[1]{\href{pmid:#1}{\path{#1}}}
\providecommand{\bibinfo}[2]{#2}
\ifx\xfnm\relax \def\xfnm[#1]{\unskip,\space#1}\fi
\bibitem[{Parapar et~al.(2023{\natexlab{a}})Parapar, Mart{\'i}n-Rodilla,
  Losada, and Crestani}]{eRisk20232}
\bibinfo{author}{J.~Parapar}, \bibinfo{author}{P.~Mart{\'i}n-Rodilla},
  \bibinfo{author}{D.~E. Losada}, \bibinfo{author}{F.~Crestani},
\newblock \bibinfo{title}{erisk 2023: Depression, pathological gambling, and
  eating disorder challenges},
\newblock in: \bibinfo{editor}{J.~Kamps}, \bibinfo{editor}{L.~Goeuriot},
  \bibinfo{editor}{F.~Crestani}, \bibinfo{editor}{M.~Maistro},
  \bibinfo{editor}{H.~Joho}, \bibinfo{editor}{B.~Davis},
  \bibinfo{editor}{C.~Gurrin}, \bibinfo{editor}{U.~Kruschwitz},
  \bibinfo{editor}{A.~Caputo} (Eds.), \bibinfo{booktitle}{Advances in
  Information Retrieval}, \bibinfo{publisher}{Springer Nature Switzerland},
  \bibinfo{address}{Cham}, \bibinfo{year}{2023}{\natexlab{a}}, pp.
  \bibinfo{pages}{585--592}.
\bibitem[{Parapar et~al.(2023{\natexlab{b}})Parapar, Rodilla, Losada, and
  Crestani}]{erisk2023}
\bibinfo{author}{J.~Parapar}, \bibinfo{author}{P.~M. Rodilla},
  \bibinfo{author}{D.~Losada}, \bibinfo{author}{F.~Crestani},
\newblock \bibinfo{title}{{Overview of eRisk 2023: Early Risk Prediction on the
  Internet. }},
\newblock in: \bibinfo{booktitle}{Experimental IR Meets Multilinguality,
  Multimodality, and Interaction. Proceedings of the 14th International
  Conference of the CLEF Association, CLEF 2023}, \bibinfo{publisher}{Springer
  International Publishing}, \bibinfo{address}{Thessaloniki, Greece},
  \bibinfo{year}{2023}{\natexlab{b}}.
\bibitem[{Conneau et~al.(2019)Conneau, Khandelwal, Goyal, Chaudhary, Wenzek,
  Guzm{\'{a}}n, Grave, Ott, Zettlemoyer, and Stoyanov}]{xlmroberta}
\bibinfo{author}{A.~Conneau}, \bibinfo{author}{K.~Khandelwal},
  \bibinfo{author}{N.~Goyal}, \bibinfo{author}{V.~Chaudhary},
  \bibinfo{author}{G.~Wenzek}, \bibinfo{author}{F.~Guzm{\'{a}}n},
  \bibinfo{author}{E.~Grave}, \bibinfo{author}{M.~Ott},
  \bibinfo{author}{L.~Zettlemoyer}, \bibinfo{author}{V.~Stoyanov},
\newblock \bibinfo{title}{Unsupervised cross-lingual representation learning at
  scale},
\newblock \bibinfo{journal}{CoRR} \bibinfo{volume}{abs/1911.02116}
  (\bibinfo{year}{2019}). \URLprefix \url{http://arxiv.org/abs/1911.02116}.
  \href{http://arxiv.org/abs/1911.02116}{{\tt arXiv:1911.02116}}.
\bibitem[{Liu et~al.(2019)Liu, Ott, Goyal, Du, Joshi, Chen, Levy, Lewis,
  Zettlemoyer, and Stoyanov}]{roberta}
\bibinfo{author}{Y.~Liu}, \bibinfo{author}{M.~Ott}, \bibinfo{author}{N.~Goyal},
  \bibinfo{author}{J.~Du}, \bibinfo{author}{M.~Joshi},
  \bibinfo{author}{D.~Chen}, \bibinfo{author}{O.~Levy},
  \bibinfo{author}{M.~Lewis}, \bibinfo{author}{L.~Zettlemoyer},
  \bibinfo{author}{V.~Stoyanov},
\newblock \bibinfo{title}{Roberta: {A} robustly optimized {BERT} pretraining
  approach},
\newblock \bibinfo{journal}{CoRR} \bibinfo{volume}{abs/1907.11692}
  (\bibinfo{year}{2019}). \URLprefix \url{http://arxiv.org/abs/1907.11692}.
\bibitem[{M{\'a}rmol-Romero et~al.(2022)M{\'a}rmol-Romero, Jim{\'e}nez-Zafra,
  Plaza-del Arco, Molina-Gonz{\'a}lez, Mart{\'\i}n-Valdivia, and
  Montejo-R{\'a}ez}]{marmol2022sinai}
\bibinfo{author}{A.~M. M{\'a}rmol-Romero}, \bibinfo{author}{S.~M.
  Jim{\'e}nez-Zafra}, \bibinfo{author}{F.~M. Plaza-del Arco},
  \bibinfo{author}{M.~D. Molina-Gonz{\'a}lez}, \bibinfo{author}{M.-T.
  Mart{\'\i}n-Valdivia}, \bibinfo{author}{A.~Montejo-R{\'a}ez},
\newblock \bibinfo{title}{Sinai at erisk@ clef 2022: Approaching early
  detection of gambling and eating disorders with natural language processing}
  (\bibinfo{year}{2022}).
\bibitem[{Pedregosa et~al.(2011)Pedregosa, Varoquaux, Gramfort, Michel,
  Thirion, Grisel, Blondel, Prettenhofer, Weiss, Dubourg et~al.}]{scikit}
\bibinfo{author}{F.~Pedregosa}, \bibinfo{author}{G.~Varoquaux},
  \bibinfo{author}{A.~Gramfort}, \bibinfo{author}{V.~Michel},
  \bibinfo{author}{B.~Thirion}, \bibinfo{author}{O.~Grisel},
  \bibinfo{author}{M.~Blondel}, \bibinfo{author}{P.~Prettenhofer},
  \bibinfo{author}{R.~Weiss}, \bibinfo{author}{V.~Dubourg}, et~al.,
\newblock \bibinfo{title}{Scikit-learn: Machine learning in python},
\newblock \bibinfo{journal}{Journal of machine learning research}
  \bibinfo{volume}{12} (\bibinfo{year}{2011}) \bibinfo{pages}{2825--2830}.
\bibitem[{Wolf et~al.(2020)Wolf, Debut, Sanh, Chaumond, Delangue, Moi, Cistac,
  Rault, Louf, Funtowicz, Davison, Shleifer, von Platen, Ma, Jernite, Plu, Xu,
  Scao, Gugger, Drame, Lhoest, and Rush}]{transformers}
\bibinfo{author}{T.~Wolf}, \bibinfo{author}{L.~Debut},
  \bibinfo{author}{V.~Sanh}, \bibinfo{author}{J.~Chaumond},
  \bibinfo{author}{C.~Delangue}, \bibinfo{author}{A.~Moi},
  \bibinfo{author}{P.~Cistac}, \bibinfo{author}{T.~Rault},
  \bibinfo{author}{R.~Louf}, \bibinfo{author}{M.~Funtowicz},
  \bibinfo{author}{J.~Davison}, \bibinfo{author}{S.~Shleifer},
  \bibinfo{author}{P.~von Platen}, \bibinfo{author}{C.~Ma},
  \bibinfo{author}{Y.~Jernite}, \bibinfo{author}{J.~Plu},
  \bibinfo{author}{C.~Xu}, \bibinfo{author}{T.~L. Scao},
  \bibinfo{author}{S.~Gugger}, \bibinfo{author}{M.~Drame},
  \bibinfo{author}{Q.~Lhoest}, \bibinfo{author}{A.~M. Rush},
\newblock \bibinfo{title}{Transformers: State-of-the-art natural language
  processing},
\newblock in: \bibinfo{booktitle}{Proceedings of the 2020 Conference on
  Empirical Methods in Natural Language Processing: System Demonstrations},
  \bibinfo{publisher}{Association for Computational Linguistics},
  \bibinfo{address}{Online}, \bibinfo{year}{2020}, pp. \bibinfo{pages}{38--45}.
  \URLprefix \url{https://aclanthology.org/2020.emnlp-demos.6}.
  \DOIprefix\doi{10.18653/v1/2020.emnlp-demos.6}.
\bibitem[{Paszke et~al.(2019)Paszke, Gross, Massa, Lerer, Bradbury, Chanan,
  Killeen, Lin, Gimelshein, Antiga, Desmaison, Kopf, Yang, DeVito, Raison,
  Tejani, Chilamkurthy, Steiner, Fang, Bai, and Chintala}]{pytorch}
\bibinfo{author}{A.~Paszke}, \bibinfo{author}{S.~Gross},
  \bibinfo{author}{F.~Massa}, \bibinfo{author}{A.~Lerer},
  \bibinfo{author}{J.~Bradbury}, \bibinfo{author}{G.~Chanan},
  \bibinfo{author}{T.~Killeen}, \bibinfo{author}{Z.~Lin},
  \bibinfo{author}{N.~Gimelshein}, \bibinfo{author}{L.~Antiga},
  \bibinfo{author}{A.~Desmaison}, \bibinfo{author}{A.~Kopf},
  \bibinfo{author}{E.~Yang}, \bibinfo{author}{Z.~DeVito},
  \bibinfo{author}{M.~Raison}, \bibinfo{author}{A.~Tejani},
  \bibinfo{author}{S.~Chilamkurthy}, \bibinfo{author}{B.~Steiner},
  \bibinfo{author}{L.~Fang}, \bibinfo{author}{J.~Bai},
  \bibinfo{author}{S.~Chintala},
\newblock \bibinfo{title}{Pytorch: An imperative style, high-performance deep
  learning library},
\newblock in: \bibinfo{booktitle}{Advances in Neural Information Processing
  Systems 32}, \bibinfo{publisher}{Curran Associates, Inc.},
  \bibinfo{year}{2019}, pp. \bibinfo{pages}{8024--8035}. \URLprefix
  \url{http://papers.neurips.cc/paper/9015-pytorch-an-imperative-style-high-performance-deep-learning-library.pdf}.

\end{thebibliography}




\end{document}